\documentclass[10pt,twocolumn,letterpaper]{article}
\usepackage[pagenumbers]{iccv}   
\usepackage{algorithm}
\usepackage{algpseudocode}
\usepackage{float}
\usepackage{subcaption}
\usepackage{stfloats}

\definecolor{iccvblue}{rgb}{0.21,0.49,0.74}
\usepackage[pagebackref,breaklinks,colorlinks,citecolor=iccvblue]{hyperref}

\newcommand{\methodname}{DiffDoctor}
\newcommand{\method}{\texttt{\methodname}\xspace}

\title{\methodname: Diagnosing Image Diffusion Models Before Treating}

\author{
    Yiyang Wang$^{1,2,*}$ \quad
    Xi Chen$^{1,2}$ \quad
    Xiaogang Xu$^{4}$ \quad
    Sihui Ji$^{1}$ \\
    Yu Liu$^{2}$ \quad
    Yujun Shen$^{3}$ \quad
    Hengshuang Zhao$^{1,\dagger}$\\[2pt]
    $^{1}$The University of Hong Kong \quad
    $^{2}$Tongyi Lab \quad
    $^{3}$Ant Group \quad
    $^{4}$Zhejiang University \\
}

\begin{document}
\maketitle

\begin{abstract}
In spite of recent progress, image diffusion models still produce artifacts.
A common solution is to leverage the feedback provided by quality assessment systems or human annotators to optimize the model, where images are generally rated in their entirety.
In this work, we believe \textbf{problem solving starts with identification}, yielding the request that the model should be aware of not only the presence of defects in an image, but  also their specific locations.
Motivated by this, we propose \method, a two-stage pipeline to assist image diffusion models in generating fewer artifacts.
Concretely, the first stage targets developing a robust artifact detector, for which we collect a dataset of over 1M flawed synthesized images and set up an efficient human-in-the-loop annotation process, incorporating a carefully designed class-balance strategy.
The learned artifact detector is then involved in the second stage to optimize the diffusion model by providing pixel-level feedback.
Extensive experiments on text-to-image diffusion models demonstrate the effectiveness of our artifact detector as well as the soundness of our diagnose-then-treat design.
\end{abstract}
\section{Introduction}
\label{sec:intro}

\begin{figure}[t]
    \centering
    \includegraphics[width=\linewidth]{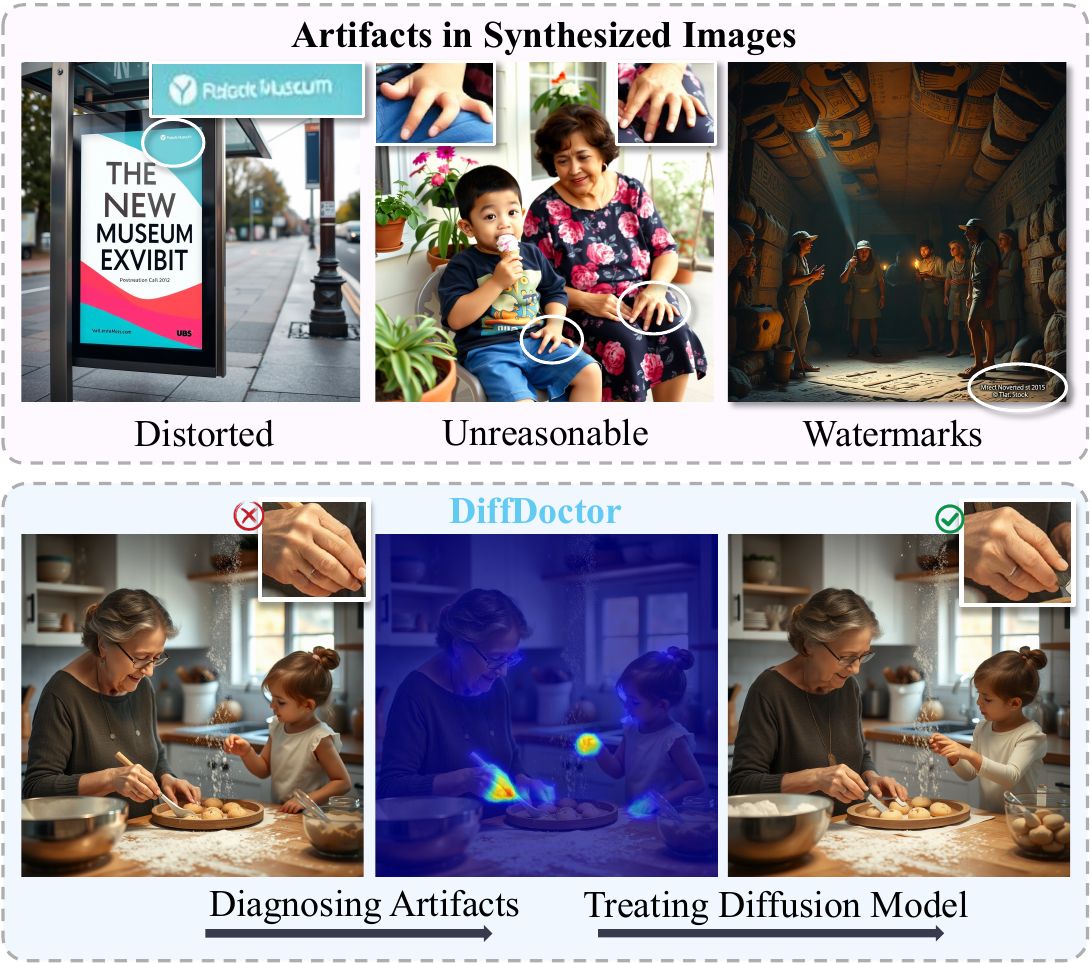}
    \vspace{-15pt}
    \caption{\textbf{Illustrations of \method.} 
    We train a robust detector to localize the artifacts~(diagnosing) and provide pixel-level feedback to optimize the diffusion model~(treating).
    After tuning on limited samples, the diffusion model generates significantly fewer artifacts on unseen prompts while maintaining the quality.
     }
    \label{fig: teaser}
    \vspace{-15pt}
\end{figure}

The advancement of image diffusion models~\cite{ho2020denoising, liu2023flow, Rombach_2022_CVPR, Flux} has made it possible to synthesize various images based on different conditions.
However, these models may still synthesize distorted, unreasonable, and unwanted content in the images, known as artifacts~\cite{liang2024rich, cao2024synartifact, zhang2023perceptual}, as shown in \cref{fig: teaser}. 
Unwanted artifacts make the outputs of the image diffusion models unstable, 
posing a significant challenge to the wider use of generative models in real-world applications.

Why do artifacts appear in synthesized images? We assert that artifacts stem from the training data's inherent noise and the model’s limited capacity. Regarding data, 
diffusion models are predominantly trained on web-mined images, which are frequently noisy~\cite{betker2023improving, kang2023noise}. 
However, standard diffusion losses~\cite{ho2020denoising, liu2023flow} do not differentiate between valuable supervisory signals and noise in the training data, leading the model to not only learn to generate vivid images but also inadvertently learn from the noise, which manifests as artifacts during image generation.
Regarding model capacity, the model’s limited capacity prevents it from fully capturing and representing complex visual details in every synthesis, occasionally resulting in artifacts.
Therefore, image diffusion models have the potential to generate correct images but lack the stability to do so reliably with each synthesis. One way to refine the pre-trained model toward more stable desired outputs is by using feedback, which can be a score describing the quality of the image, and current methods attempt to maximize such score.
However, feedback of these methods is generally an image-level score~\cite{xu2024imagereward, black2024training, fan2024reinforcement, zhang2024large, clark2024directly, prabhudesai2023aligning, cao2024synartifact,kirstain2023pick,wu2023human} to each image, or pairwise comparisons~\cite{wallace2024diffusion},
overlooking fine-grained pixel-level information — artifacts are sparse within the image.
This analysis shapes the motivation for our paper: \textit{Problem-solving starts with identification. We should know where artifacts are before tuning the diffusion model.} 

Therefore, we present \method, 
which tunes diffusion models in a diagnose-then-treat diagram.
Diagnosing means identifying the artifact locations in synthesized images, which requires an artifact detector to predict artifact confidence.
Treating means supervising the diffusion model with the artifact detector, where we minimize the per-pixel artifact confidence of synthesized images, aiming to reduce artifacts in future synthesis. Illustrations are in \cref{fig: teaser}.

To train a robust artifact detector, it is essential to collect comprehensive data. 
RichHF~\cite{liang2024rich} and PAL4VST~\cite{zhang2023perceptual} provide dense artifact annotations.
However, we identify severe data-imbalance issues -- overwhelming positive samples on specific categories (\textit{e.g.,} hands are usually wrong).
Training on them leads to a high false positive rate of the artifact detector in these categories (\textit{e.g.,} predict artifacts for every hand even when correct).
Therefore, we develop a class-balancing strategy to collect additional data to balance negative and positive artifact samples. Using this strategy, we established a human-in-the-loop pipeline to collect and label images synthesized by various diffusion models. Hard cases are meticulously labeled by human annotators to ensure accurate target signals, while easier cases are auto-labeled~\cite{depth_anything_v1} by the trained artifact detector using a specially designed augmentation strategy. Together, these efforts result in a robust artifact detector trained on 1M+ samples.

After training a robust artifact detector, we treat the diffusion model with its feedback.
The detector predicts confidence maps, where each pixel’s value represents the confidence (or probability) of artifact presence, which we aim to minimize.
Specifically, we prompt the to-optimize diffusion model to synthesize images, 
detect artifact maps using the artifact detector, and directly back-propagate the gradients by minimizing the artifact confidence of each pixel of the synthesized images back to the diffusion model.
This method leverages the pixel-level information in artifact-prone areas, allowing precise treating (or penalizing) of the diffusion model to make the model try to avoid these artifacts in the future.
At the same time,  this training paradigm seamlessly accommodates the original diffusion loss, which we leverage as a regularization to alleviate model collapse. 

To the best of our knowledge, \method is the first approach to use pixel-level feedback for tuning image diffusion models. Tuned on limited prompts, the diffusion model generates fewer artifacts of similar types on unseen prompts while maintaining the quality.
It's also applicable to other methods such as DreamBooth~\cite{ruiz2023dreambooth}.
Our contributions are summarized as follows:
\begin{itemize}
    \item We propose \method, the first approach to use pixel-level feedback to tune image diffusion models to reduce artifacts using a diagnose-then-treat design.
    \item To diagnose artifacts, we train a robust artifact detector by using a careful class-balance strategy and scaling up the data with human-in-the-loop.
    \item To treat artifacts, we explore different solutions to leverage the artifact map to provide per-pixel feedback and an optional regularization term to alleviate early collapse.
\end{itemize}
    
\section{Related Work}
\label{sec:related}

\noindent\textbf{Detect abnormality in images.}
Our work focuses on detecting and thereby mitigating artifacts in \textit{synthesized} images of diffusion models, which may be confused with detecting GenAI-generated or altered images~\cite{yu2024diffforensics, xu2024fakeshield,zhang2022perceptual,wang2019detecting}. These works aim to identify whether AI has generated or manipulated a real image, rather than assessing the presence of artifacts within synthesized images. 
Now for the artifacts in synthesized images.
SynArtifact~\cite{cao2024synartifact} uses VLMs to classify artifacts or to locate them using bounding boxes.
PAL4VST~\cite{zhang2023perceptual} and RichHF~\cite{liang2024rich} concurrently provide the first open-sourced data with annotated artifact areas for synthesized images. However, these artifact annotations suffer from limitations of imbalance and coarse granularity, which can significantly compromise the effectiveness of an artifact detector trained on them. In this work, we carefully design a class-balance strategy and set up a human-in-the-loop annotation process, resulting in a robust artifact detector trained on 1M+ data samples.

\noindent\textbf{Guide diffusion models with feedback.}
Recently, inspired by RLHF for LLMs~\cite{ouyang2022training},
ImageReward~\cite{xu2024imagereward} introduces an image reward model to approximate human feedback with image-level scores and optimizes diffusion models via ReFL to maximize these scores. 
DDPO and DPOK~\cite{black2024training,fan2024reinforcement} apply reinforcement learning (RL)~\cite{schulman2017proximal} to make diffusion models maximize a global score (e.g., aesthetic score). 
Zhang et al.~\cite{zhang2024large} is the first to apply RL at scale, training diffusion models with millions of prompts and multiple reward objectives.
SynArtifact~\cite{cao2024synartifact} globally classifies artifacts as scores and applies RL to tune diffusion models.
Diffusion-DPO~\cite{wallace2024diffusion} applies direct preference optimization~\cite{rafailov2024direct} to diffusion models, transforming the RL problem into a classification objective between winning and losing samples.
AlignProp~\cite{prabhudesai2023aligning} and DRaFT~\cite{clark2024directly} directly optimize diffusion models by back-propagating from differentiable reward models to maximize the image-level score. 
However, these models rely on image-level ratings or comparisons, tuning the model to maximize overall scores and overlooking the fine-grained information in each pixel.
PAL4VST~\cite{zhang2023perceptual} and RichHF~\cite{liang2024rich} are the first to leverage fine-grained artifact annotations. However, they don't use these fine-grained annotations to fine-tune diffusion models; instead, they simply generate a batch of images to rank based on artifact confidence or mask out problematic areas for inpainting.
In this paper, we instead diagnose pixel-level artifacts to tune (treat) diffusion models, enabling them to try to avoid generating artifacts in the future. To the best of our knowledge, \method is the first approach to utilizing pixel-level feedback for tuning diffusion models.
\begin{figure*}[t]
    \centering
    \resizebox{\linewidth}{!}{
    \includegraphics{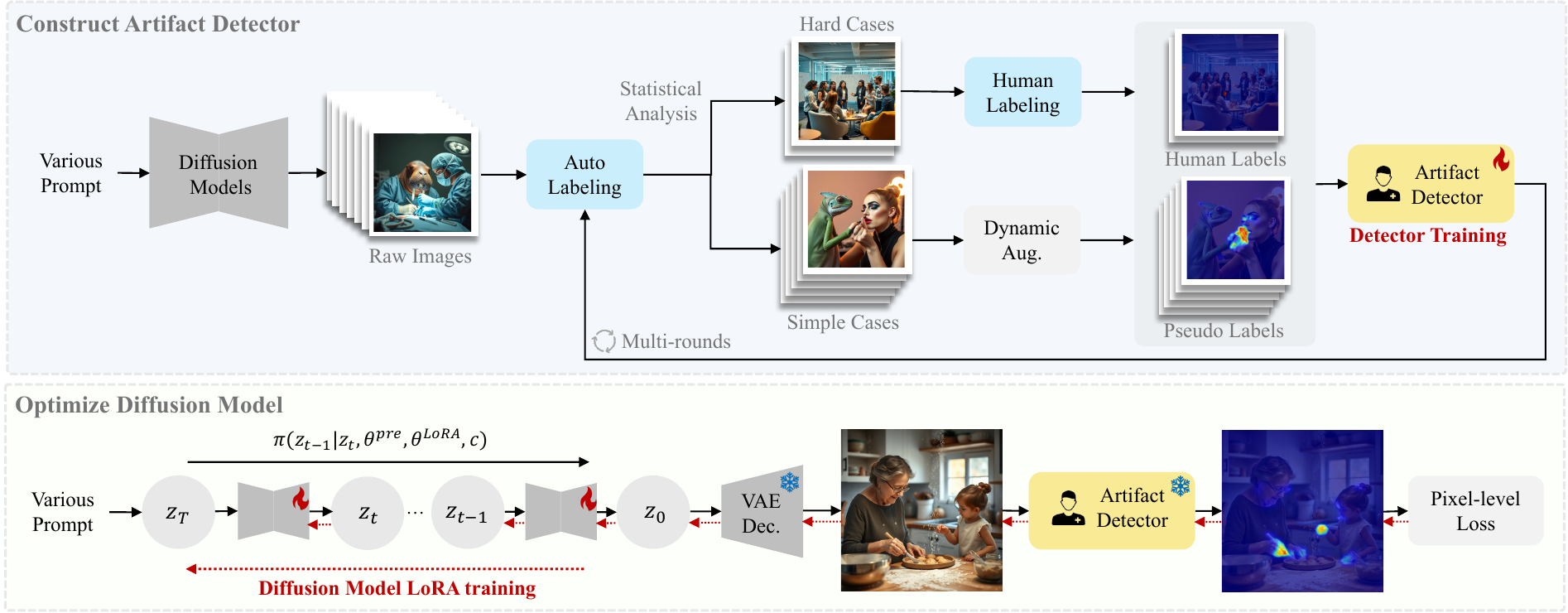}
    }
    \vspace{-20pt}
    \caption{\textbf{Pipeline of \method.} 
    The first part shows the training of an artifact detector -- the doctor. Starting with the initial dataset, the artifact detector is trained in a humans-in-a-loop manner.
    The second part shows our diagnose-then-treat design, where the patient -- a trainable diffusion model, is prompted to synthesize images. Then the frozen artifact detector diagnoses its result by predicting the artifact maps, on which it treats the patient by minimizing the per-pixel artifact confidence to back-propagate to the diffusion model.
    }
    
    \label{fig: pipeline}
    \vspace{-5mm}
    \end{figure*}
\section{Method}

\subsection{Preliminaries}
\label{sec: Preliminaries}
\noindent\textbf{Artifact definition.}
We define artifacts as three types: shape distortions (\textit{e.g.,} distorted hands, faces, words), unreasonable content (\textit{e.g.,} extra limbs, more-or-fewer fingers), and watermarks. We do not consider artifacts requiring complex reasoning to find (\textit{e.g.} floating objects without physical support).
Examples of artifacts are in \cref{fig: teaser}.

\noindent\textbf{Artifact detection.} 
We define artifact areas in synthesized images as confidence maps (artifact maps), transforming artifact detection problems into a binary segmentation task, where each pixel represents a confidence value from 0 to 1, indicating the likelihood of the pixel being part of an artifact.
Then the training process of the artifact detector is to minimize the mean squared error (MSE) loss between the predicted artifact maps and the ground truth artifact maps:
$
    \mathcal{L}_{\text{AD}}(\theta) = \frac{1}{N} \sum_{i=1}^{N} \left\| \hat{C}_\theta(x_i) - C(x_i) \right\|_2^2,
$
where \(N\) is the number of samples in a batch, \(x_i\) is the \(i\)-th synthesized image, \(C(x_i)\) is the ground truth artifact map of \(x_i\),
and \(\hat{C}_\theta(x_i)\) is the predicted artifact map of \(x_i\).

\subsection{Construct Artifact Detector}
\label{sec: AD}

We find an imbalance of current artifact annotations compromising the artifact detector, so we design a class balance strategy with human-in-the-loop to train a robust artifact detector on 1M+ balanced data samples, a mix of active learning and semi-supervised learning. The pipeline is in \cref{fig: pipeline}: We generate numerous images using various diffusion models, selecting hard cases for human labeling through statistical analysis, while simpler cases are automatically labeled by the updated artifact detector, in multi-rounds.

\noindent\textbf{Data-imbalance issue.} 
Existing datasets with pixel-level artifact annotations \cite{zhang2023perceptual,liang2024rich} provide synthesized images with artifact maps, offering a training set of 25K samples and establishing a foundation for artifact detector construction.
However, we identify a limitation due to data imbalance in previous data, which seriously compromises the performance of the artifact detector.
Specifically, the negative (artifact-free) and positive (artifact-prone) samples are significantly unbalanced in certain categories.
For instance, human-centered images synthesized with inferior models usually have distorted hands or faces (like abnormal fingers), which dominate the training set in previous annotations. These overwhelming positive samples cause the artifact detector to learn a shortcut where it always predicts high confidence in these regions, even if they are not artifacts.
This issue goes unnoticed because previous testing set is also imbalanced, which hides the high false positive rates problem during evaluation.

\noindent\textbf{Balance data distribution.} 
We first incorporate high-quality real photos as negative samples to balance the distribution, as they are artifact-free, easy to obtain, and don't require labeling. 
After adding real images, the false positive rate of the artifact detector on real photos reduces significantly.
However, this data still has limitations in generalizing to synthesized images due to the domain gap. Therefore,  we further incorporate synthesized images. 

Different from label-free real photos, synthesized images require dense annotations for artifact areas.
Given the expensive cost of extensively labeling artifact areas,
we only label the most informative and hard cases that are challenging for artifact detectors to learn, following the idea of active learning.
The imbalance problem arises because images of certain categories (\textit{e.g.}, humans) are hard for inferior diffusion models to generate, resulting in synthesized images full of artifacts in previous datasets. 
These images are overwhelmed with positive artifact signals, making it hard for the artifact detector to discriminate.
We find these hard categories by using MLLM~\cite{bai2023qwen} to assign categories to images, calculate the aggregated artifact confidence of each category, and find the categories with abnormally high or low confidence.
Then, we use an LLM~\cite{qwen} to generate various prompts for these categories and synthesize images using SOTA text-to-image models. Compared with inferior models, they can synthesize more artifact-free images, serving as the challenging cases that are balanced with positive and negative artifact signals.
From them, we select a subset of 2k hard cases by thresholding the images detected with extremely high or low artifact confidence on these categories,
which we then manually label to provide precise guidance signals for joint training.

\noindent\textbf{Scale-up with unlabeled images.} 
To further enhance the generalization ability and robustness of the artifact detector, we scale up the training data by predicting pseudo-labels for the images not selected as hard cases, following studies in semi-supervised learning~\cite{depth_anything_v1,depth_anything_v2,yang2022st++,yang2023revisiting}. 
Illustrated in the first part of \cref{fig: pipeline}, we jointly train the artifact detector on both true labels and pseudo labels with humans in the loop, resulting in 1M+ data samples involved in training.
We also design a dynamic augmentation strategy specifically for artifact detection to perturb the pseudo labels.

The dynamic augmentation strategy shrinks images with a predicted maximum artifact confidence value below a certain threshold to smaller sizes and then pads them back to original sizes. 
This is because small complex regions (\textit{e.g.}, a face of $60\times70$) are more likely to have artifacts. Whereas, large areas (\textit{e.g.}, a face of $300\times350$) are more likely to be correct.
Therefore, these shrunk images are more likely to contain small complex regions that are artifact-free, which helps to balance the distribution.
Apart from this, we by default apply strong augmentations for pseudo labels following previous semi-supervised learning studies~\cite{yang2023revisiting}.

\noindent\textbf{Robust artifact detector.} 
The above content demonstrates how we balance and scale up the data.
In terms of model backbone selection, we train our artifact detector using SegFormer-b5~\cite{xie2021segformer} as the backbone.
We directly apply a sigmoid function after the logit outputs to get confidence values for each pixel, 
which is output as the artifact map. 

\subsection{Diagnose-then-treat Diffusion Models}
The artifact detector enables us to diagnose artifacts within synthesized images, serving as valuable dense supervisory signals for artifacts. Although RichHF~\cite{liang2024rich} and PAL4VST~\cite{zhang2023perceptual} also attempt to locate artifacts in synthesized images, they just use this information to inpaint problematic areas or to rank batches of images. How to leverage these pixel-level artifacts to supervise diffusion models remains underexplored. Therefore, we present pixel-aware treating, which tunes the diffusion model with the supervision from the artifact detector as
shown in \cref{fig: pipeline}.

\label{sec:diffusion}
\noindent\textbf{Pixel-aware treating.}
The artifact detector is differentiable, so we can directly back-propagate gradients from detected artifact maps of synthesized images to the diffusion model through the artifact detector. 
We aim to minimize all confidence  of the artifact map of the synthesized image (\textit{i.e.,} punish high artifact confidence), leading to the following pixel-level loss for supervising the diffusion model:

\begin{align}
    \mathcal{L}_\text{pixel}(\theta) = \frac{1}{N_{\text{aggr}}}\sum_{i,j}^{h,w} M\circ C(\pi_\theta(z_T))[i,j] , 
\end{align}
where \(\pi_\theta(z_T)\) represents the image sampled from the diffusion policy model \(\pi_\theta\) starting from the standard Gaussian noise \(z_T\).
Note that the diffusion model performs the denoising process with gradient tracking.
\(C(\pi_\theta(z_T))\) is the artifact map of this synthesized image
diagnosed by the artifact detector,
\(h,w\) are the height and width, 
and \(i,j\) traverse all pixel coordinates.
\(M\) represents a threshold mask deciding the pixel involved in the calculation, and \(N_{\text{aggr}}\) is the number of calculated pixels.
This loss suppresses the diffusion model for generating high artifact confidence areas by supervising the diffusion model on pixel-level artifact signals from the artifact detector.
Illustration is shown in \cref{fig: pipeline}

Following studies of direct reward fine-tuning~\cite{clark2024directly, prabhudesai2023aligning} which also directly back-propagating diffusion models but with global reward models when doing full-chain sampling,
we truncate the gradients in the last few steps in the denoising chain to save memory.

\noindent\textbf{Compatibility with diffusion loss.}
Pixel-aware treating is also compatible with standard diffusion losses, which can also help delay the model collapse.
Regarding collapse, overly treating the model will lead to quality degradation (model collapse) similar to the reward hacking problem, as discussed in \cref{sec: ablation}. We can avoid this problem by using early stopping; 
however, we still consider minimizing KL regularization to constrain updated models from the real image distribution to mitigate collapse:
\(\mathbb{E}_{p(y)}[\text{KL}(p_\theta(x|y)||p_\text{real}(x|y))]\),
which can be transformed into a rectified flow loss~\cite{liu2023flow, esser2403scaling}:
\begin{align}
\mathcal{L}_\text{offline}(\theta) = || (z_T-z_0) - v_\theta(z_t,t)  ||.
\end{align}

We call it an offline regularization term and derive the final loss with the regularization:
    $ \mathcal{L} = \mathcal{L_\text{pixel}} +  \gamma \mathcal{L}_\text{offline},
$
where we empirically choose \(\gamma = 0.25\). 
In this case, \(\mathcal{L_\text{pixel}}\) is seamlessly fused with a diffusion loss, and only one diffusion model is required to load, which just slightly increases the training overhead for treating.
The pseudocode with the offline regularization term is shown in \cref{alg: reflection}, where ``no grad'' means disabling gradient tracking and ``with grad'' performs gradient tracking. 

\begin{algorithm}[t]
    \small
    \caption{\method w/ Offline Regularization}
    \label{alg: reflection}
    \begin{algorithmic}[1]
    \State \textbf{Prompt Set:} $\mathcal{C} = \{c_1, c_2, \dots, c_n\}$
    \State \textbf{Real Image Set:} $\mathcal{D} = \{I_1,I_2, \dots, I_m\}$
    \State \textbf{Input:} Base diffusion model with pre-trained parameters $\theta_0$, artifact detector $\mathcal{AD}$, regularization scale $\gamma$
    \State \textbf{Initialization:} Noise scheduler timestep number $T$, gradient truncation time step $t_{\text{tr}}$
    \While{not converged}
        \For{$c_j \in \mathcal{C}$}
            \State $z_T \sim \mathcal{N}(0, \mathbf{I})$
            \For{$i = T, \dots, t_{\text{tr}} + 1$}
                \State \textbf{no grad:} $z_{i-1} \gets p_\theta(z_i |i;c_j)$
            \EndFor
            \For{$i = t_{\text{tr}}, \dots, 1$}
                \State \textbf{with grad:} $z_{i-1} \gets p_\theta(z_i|i;c_j)$
            \EndFor

            \State $x \gets \text{VAEdec}(z_0)$ \Comment{Decode the image}
            \State $C(x) \gets \mathcal{AD}(x)$ \Comment{Get the artifact map}
            \State $\mathcal{L}_\text{pixel} \xleftarrow{\text{aggregate}} C(x)$ \Comment{Pixel-level loss}
            
            \State $I_{k} \gets \text{select}(\mathcal{D})$ \Comment{Select a real image}
            \State $\mathcal{L}_\text{offline} \xleftarrow{\text{diffusion}} I_{k}$ \Comment{standard diffusion loss}
            
            \State $\mathcal{L} \gets \mathcal{L}_\text{pixel} + \gamma \mathcal{L}_\text{offline}$ \Comment{Final loss}
            \State $p_{\theta_{i+1}} \gets p_{\theta_{i}}$ \Comment{Update the diffusion model}
        \EndFor
    \EndWhile
    \end{algorithmic}
\end{algorithm}
\section{Experiments}
\label{sec: Experiments}

\subsection{Experiment Settings}
\label{sec: settings}
\noindent\textbf{Artifact detector benchmark.} 
RichHF has a testing set, 
but their annotations are circles with fixed radii. These fixed-radius circles usually cover non-artifact areas redundantly or fail to cover the entire artifact, leading to inaccurate metric values.
Therefore, we do not use this benchmark and construct a benchmark of 771 images. It contains synthesized images covering hard cases challenging for artifact detectors with fine-grained artifact labels and real photos from COCO~\cite{cocodataset} testing set to measure the false positive rates on real photos.
For the metrics, we use MSE to measure accuracy.
To measure the false negative rates, 
we utilize the mean KL divergence:
\(\text{KL}(P||Q) = \frac{1}{N}\sum_{\text{pixels}}[p(x)\log \frac{p(x)}{q(x)}]\)
where \(p(x)\) and \(q(x)\) are the ground truth and predicted confidence for pixel \(x\), \(N\) is total pixel number.
Relatively, 
\(\text{KL}(1-P||1-Q)\) 
is effective in measuring false positive rates, and we denote this metric as KL(1-) in the following sections.

\noindent\textbf{Model treating dataset.} 
The training set and the benchmark for treating only contain text prompts without training images.
We prompt Qwen~\cite{qwen} to synthesize 3100 complex prompts describing complicated and various scenarios mainly concerning human activities, animals, and words in life to challenge the diffusion models. We randomly break them into a training set of 3000 prompts, and a benchmark of 100 prompts.
For optional offline regularization, we use LAION-Aesthetics V2 6.5+ Dataset~\cite{schuhmann2022laion}.

\noindent\textbf{Model treating metrics.} 
We further emphasize that the training set for treating \textit{only contains text prompts without training images}. Therefore, it's NOT feasible to apply metrics that measure distribution distances between synthesized images and reference images, such as FID and CLIP-I score. Thus, we apply metrics that depend only on the synthesized images and text prompts, including the ImageReward score and CLIP-T score. Furthermore, we threshold max artifact confidences of images(0.5) to measure whether an image contains artifacts, based on which we calculate the ``mean artifact frequencies" of the test set.
We also conducted user studies with 24 users. We offer them pairs of images on the same diffusion model before and after treating and ask them to select a winning image based on general image quality and artifact presence, respectively. Then, we calculate the winning rate before and after treating.

\noindent \textbf{Implementation details.}
We conduct most experiments o FLUX.1 Schnell~\cite{Flux} with inference step as 4. We also apply treating on latent diffusion models such as SDXL~\cite{podell2024sdxl} and Kolors~\cite{kolors} with inference steps as 20.
We use a learning rate of \(1e-4\).
To save memory, we choose the gradient truncation timestep \(t_{tr}\) to be the last 25\% of inference step and only train LoRA~\cite{hu2022lora} layers (LoRA rank is 16). We use the best artifact detector (+ pseudo 1M) for training.

\subsection{Ablation Studies}
\label{sec: ablation}

\begin{figure}
    \centering
    \includegraphics[width=\linewidth]{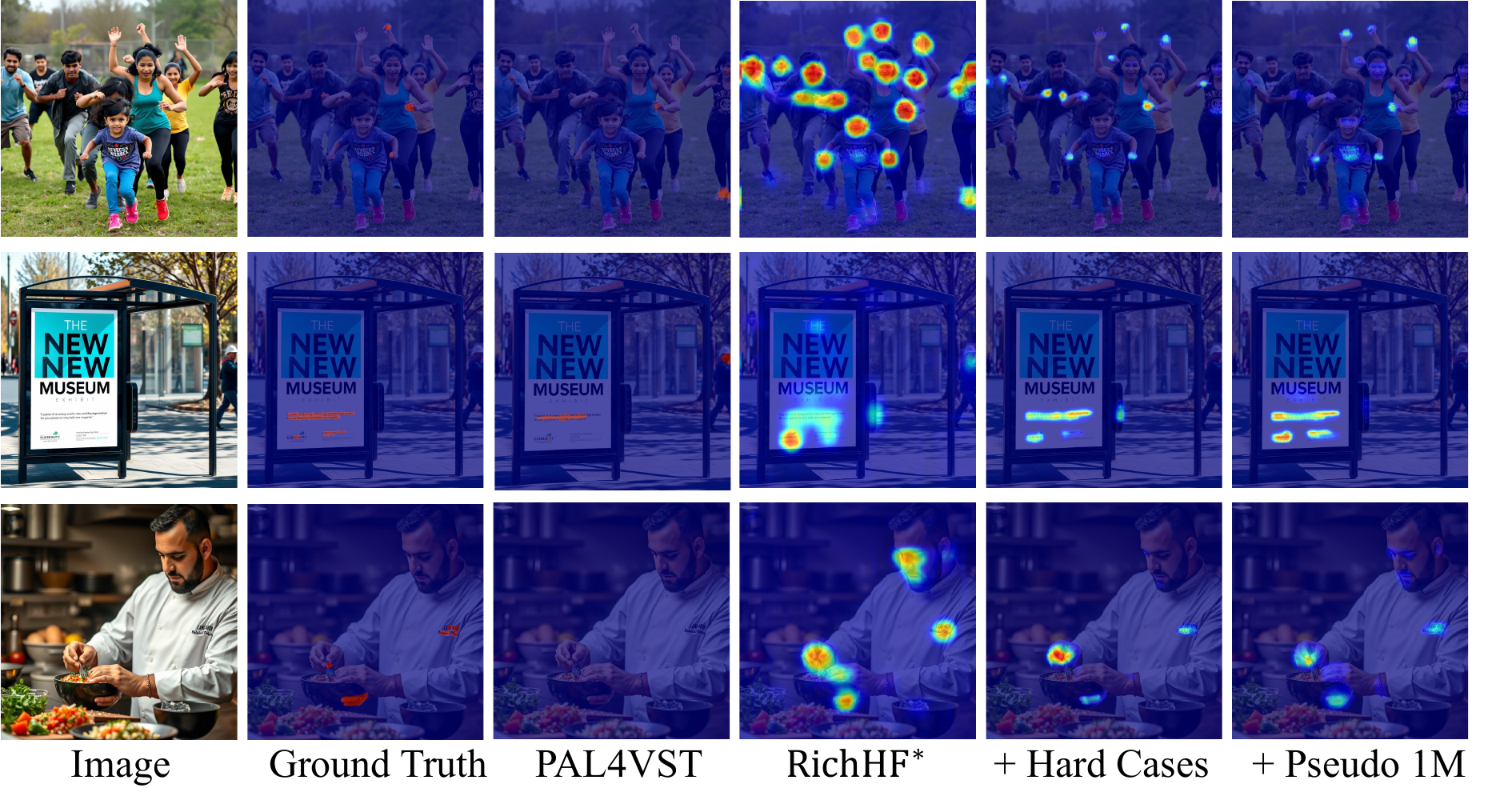}
    \vspace{-25pt}
\caption{\textbf{Qualitative ablation study of artifact detectors.}
  We visualize the artifact maps predicted by the artifact detector on our hard benchmark by headmaps.
  }
  \vspace{-3mm}
  \label{fig: AD ablation}
\end{figure}

\begin{table}[t]
    \LARGE
     \caption{\textbf{Quantitative comparisons and ablations of artifact detectors.} 
  Data are in percentages with percentage signs (\%) omitted.
  We measure
 on \textbf{our} constructed benchmark and \textbf{real} photos.
  }
  \vspace{-3mm}
    \resizebox{!}{1.15cm}{
    \begin{tabular*}{2.4\linewidth}{lcccccl}
      \toprule
      Method 
      & MSE Ours$\downarrow$ & KL Ours$\downarrow$ & KL(1-) Ours$\downarrow$ 
      & MSE real$\downarrow$ & KL(1-) real$\downarrow$ 
      \\
      \midrule
      PAL4VST  & 0.480 & 5.053 &  2.394 & 0.591 & 5.740 \\
      $\text{RichHF}^*$  & 1.601 & 1.059 & 7.044 & 0.979 & 6.082\\
      \midrule
      + real photos & 1.167 & 1.111 & 4.803 & 0.029 & 1.558 \\
      + hard cases & 0.504 & \textbf{0.981} & 2.983 & 0.003 & 0.458 \\
      + pseudo 1M & \textbf{0.337} & 1.004 & \textbf{2.231} & \textbf{0.002} & \textbf{0.371} \\
    \bottomrule
    \end{tabular*}
   }
  
  \vspace{-6mm}
  \label{tab: ad_ablation}
  \end{table}

\noindent\textbf{Artifact detector.}
We include \textbf{comparisons} due to limited space.
The RichHF artifact detector is not open-sourced, so we implement it as $\text{RichHF}^*$, also the baseline for ablations. 

When trained solely on previous artifact annotations ($\text{RichHF}^*$), the model shows high false positive rates (KL(1-)) and high MSE in our benchmark and real photos in \cref{tab: ad_ablation} (also visualized in \cref{fig: AD ablation}), as it predicts high confidence in almost all areas of faces or limbs. With real photos, the false positive rate is reduced on real photos but still high for our benchmark.
Incorporating hard cases into training further improves the model's MSE and reduces false positive rates. A great enhancement is shown in \cref{fig: AD ablation}, where adding hard cases significantly mitigates the shortcut problem, preventing the model from over-detecting faces and limbs and directing its focus more accurately on true artifacts.
Adding pseudo-labeled images further reduces false positive rates and MSE on our benchmark and real photos, with performance improving as more pseudo-labels are included. 
Using pseudo-labels will slightly increase the KL values (false negative rates). We hypothesize that this is due to the increasing negative signals in pseudo labels, causing the model to become slightly more conservative in predicting high confidence. 
However, given the significant improvements in accuracy and false positive rates that cover the slight degradation of false negative rates, we choose the model trained on 1M pseudo labels as the best artifact detector used for the following experiments for treating.
Our method also significantly outperforms PAL4VST on all metrics, especially on the false negative (KL) metric.

\begin{figure}[!t]
    \centering
    \includegraphics[width=\linewidth]{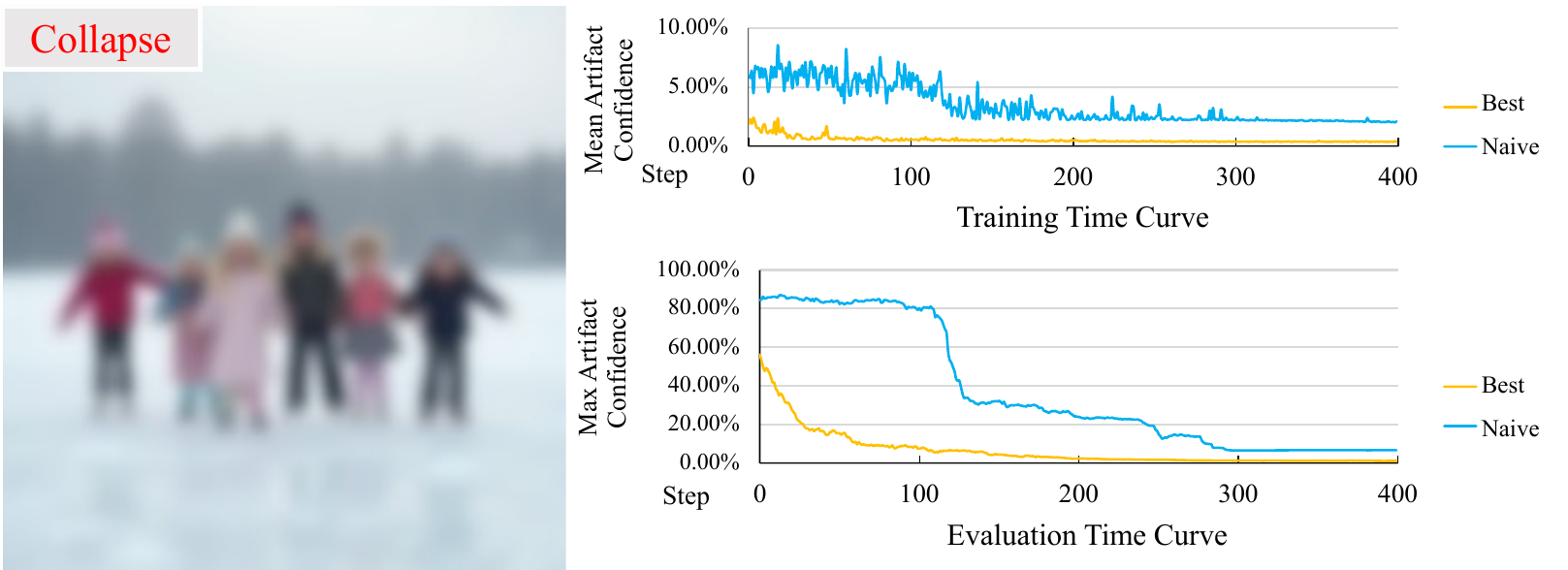}
    \vspace{-20pt}
    \caption{\textbf{Mode collapse due to naive artifact detector.} The treating process collapses into blurriness. We further visualize the training and evaluation time artifact confidence curve.
    }
    \label{fig:collapse_bad_ad}
    \vspace{-4mm}
\end{figure}

\begin{figure}[htbp]
  \centering
      \includegraphics[width=\linewidth]{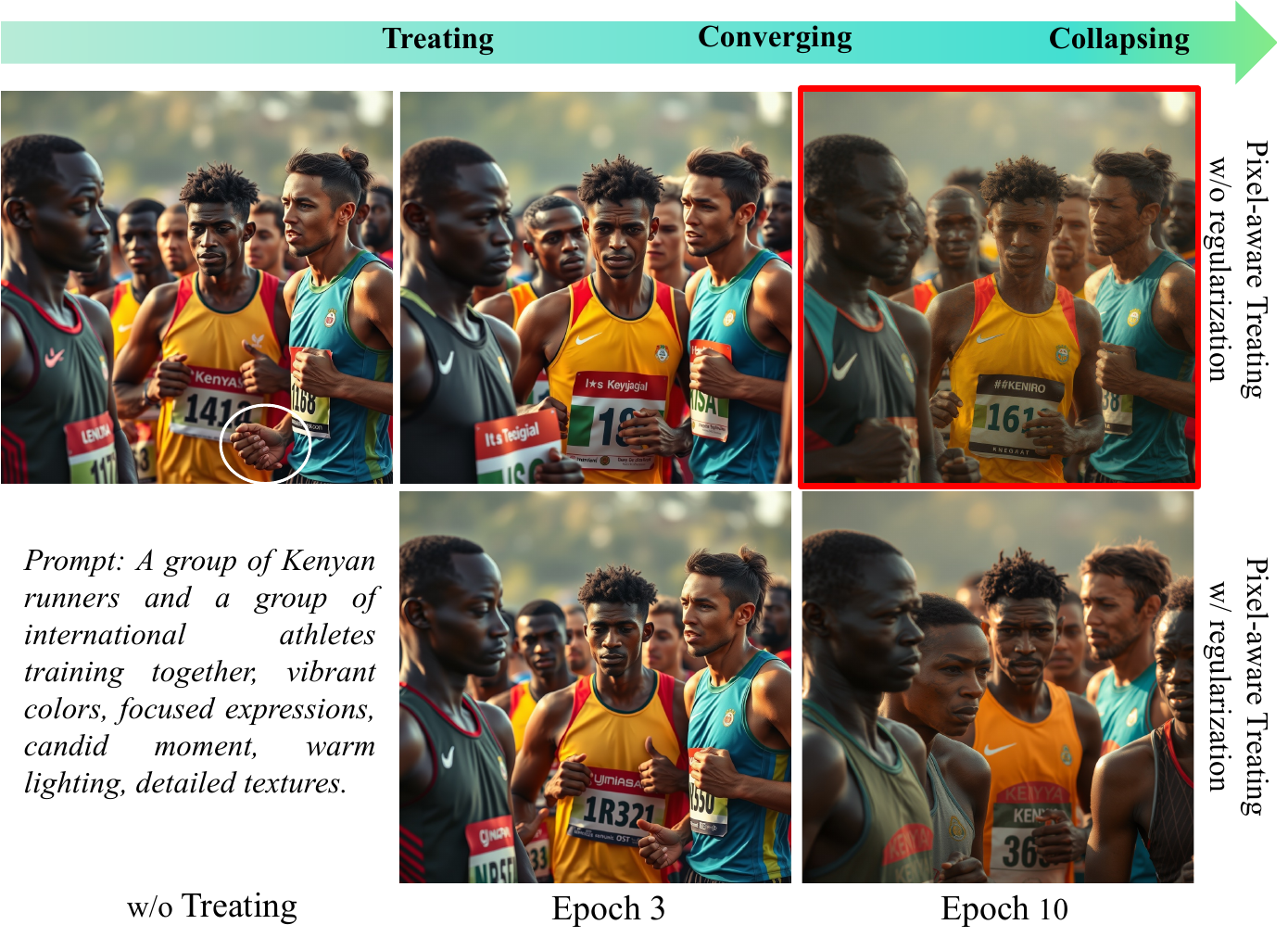}
\vspace{-8.5mm}
  \caption{\textbf{Qualitative ablation on offline regularization. }
  These images are synthesized by FLUX.1 during treating with or without offline regularization in evaluation time. We frame the synthesized image in the model collapse stage with a red box.
  }
  \vspace{-4mm}
  \label{fig: spectrum}
\end{figure}

\begin{table}[!t]
      \centering
      \caption{\textbf{Quantitative ablations of treating}. Ablations on mask selection and artifact detector selection when treating FLUX.1.
     ``All" means using all pixels in the pixel-level loss.
    }
    \vspace{-3mm}
    \resizebox{!}{1.7cm}{
    \begin{tabular}{lccl}
    \toprule
    Method  & ImageReward $\uparrow$ & CLIP-T~($\uparrow$)\\
    \midrule
    FLUX.1 (Baseline) & 1.179$\pm$0.019 & 35.463$\pm$0.047 \\
    \midrule
    All + Best Detector & 1.161$\pm$0.028 & 35.278$\pm$0.051 \\
    Max + Best Detector  & 1.159$\pm$0.043 & 35.510$\pm$0.023 \\
    \textbf{Threshold + Best Detector} & 1.183$\pm$0.024 & 35.611$\pm$0.036 \\
    \midrule
    All + Naive Detector & -0.898$\pm$0.068 & 28.960$\pm$0.028 \\
    Max + Naive Detector & -0.902$\pm$0.038 & 28.474$\pm$0.017 \\
    Threshold + Naive Detector & -0.884$\pm$0.009 & 28.431$\pm$0.014 \\
    \bottomrule
    \end{tabular}
    }
     
    \label{tab: quan_ablation}
     \vspace{-6mm}
    \end{table}

\begin{table}[t]
      \centering
      \caption{\textbf{Quantitative ablations on offline regularization}. 
     We experiment on FLUX.1 on using diffusion loss as regularization.
    }
    \vspace{-3mm}
    \resizebox{!}{1.2cm}{
    \begin{tabular}{lccl}
    \toprule
    Method  & ImageReward $\uparrow$ & CLIP-T~($\uparrow$)\\
    \midrule
    FLUX.1 (Baseline) & 1.179$\pm$0.019 & 35.463$\pm$0.047 \\
     w/o Regularization, Epoch 3 &  1.183$\pm$0.024 & 35.611$\pm$0.036 \\
     w/ Regularization, Epoch 3& 1.167$\pm$0.027 & 35.707$\pm$0.043 \\
     w/o Regularization, Epoch 10&  0.904$\pm$0.065 & 34.952$\pm$0.062 \\
     w/ Regularization, Epoch 10& 1.180$\pm$0.023 & 35.842$\pm$0.041 \\
    \bottomrule
    \end{tabular}
    }
    \label{tab: quan_ablation_diff}
     \vspace{-6mm}
    \end{table}

\noindent\textbf{Mask selection for treating.}
We also ablate on the threshold mask \(M\) and \(N_{\text{aggr}}\) in  \(\mathcal{L}_\text{pixel}\) during treating.
In \cref{tab: quan_ablation}, we use FLUX.1 to compare the approach of selecting all pixels (all), the maximum artifact confidence pixel (max), and pixels with confidence greater than 0.1 (threshold). Note that the artifact frequency metric is not feasible here, for the artifact confidences are all optimized to low values. Therefore, we mainly inspect the image quality. Thresholding the pixels gains the best results, showing that a finer control of pixel-feedback instead of simply using all pixels or one pixel will have better impacts on image quality.

\noindent\textbf{Artifact detector selection for treating.}
We verify that we must train a robust artifact detector not having high false positive errors for downstream treating.
We choose \(\text{RichHF}^*\) as the naive artifact detector. In \cref{fig:collapse_bad_ad}, its starting confidence is higher, for it demonstrates high false positive rates. 
During treating, its confidence curves in training and evaluation time both first oscillate and then drop sharply. The synthesized images collapse into blurriness after this sharp degradation. We hypothesize that the over-high false positive rates cause the diffusion model to receive incorrect high punishment for almost every synthesis, driving it to this shortcut where the model collapses seriously.
For the best artifact detector, the confidence drops normally without sharp degradation. 
Quantitative ablations on detector selection in \cref{tab: quan_ablation}  show that, using the naive detector, after the sharp degradation, the ImageReward and CLIP-T scores drop significantly due to the image quality degradation.

\noindent\textbf{Offline regularization for treating.}
We study the impacts of treating w/ or w/o offline regularization in \cref{fig: spectrum} and \cref{tab: quan_ablation_diff} on FLUX.1. 
Before treating, the model generates obvious artifacts. 
After 3 epochs, artifacts are both reduced.
For w/o regularization, the layout of the image changes slightly when trained for epochs. But in epoch 10, it synthesizes overly smooth patterns that we call collapse.
W/ regularization is more stable against collapse when trained for 10 epochs, and the image layout shifts more due to the additional supervised signals of the diffusion loss.
These indicate that model collapse can be prevented by early stopping.
Using regularization can \textit{slow} the collapse and may shift the output distribution (image layouts) more significantly (not considered a drawback).
\cref{tab: quan_ablation_diff} shows that results using early stopping are comparable in CLIP-T, while w/o regularization is better in ImageReward.
But after 10 epochs, w/o regularization wins.
Therefore, using offline regularization depends on whether we want to trade off more image layout changes with longer training time.

\label{sec: DMexperiments}

\begin{figure*}[!t]
  \centering
  \small
  \includegraphics{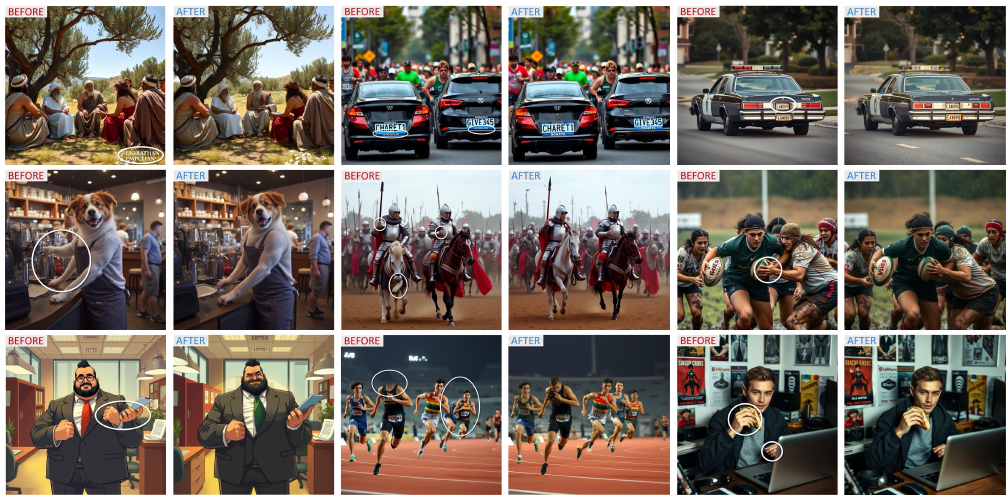}
  \vspace{-8pt}
  \caption{\textbf{\method on FLUX.1}. 
  All images are synthesized based on randomly generated unseen prompts not involved in training, and on the same seeds for the corresponding images before and after \method. After treating, artifacts in images are reduced, but the content and layouts of the images are almost unchanged, demonstrating the effectiveness of pixel-aware treating.
  }
  
  \label{fig: reflection_exp_FLUX}
  \vspace{-14pt}
  \end{figure*}

\begin{figure}[t]
\centering 
\includegraphics[width=1.0\linewidth]{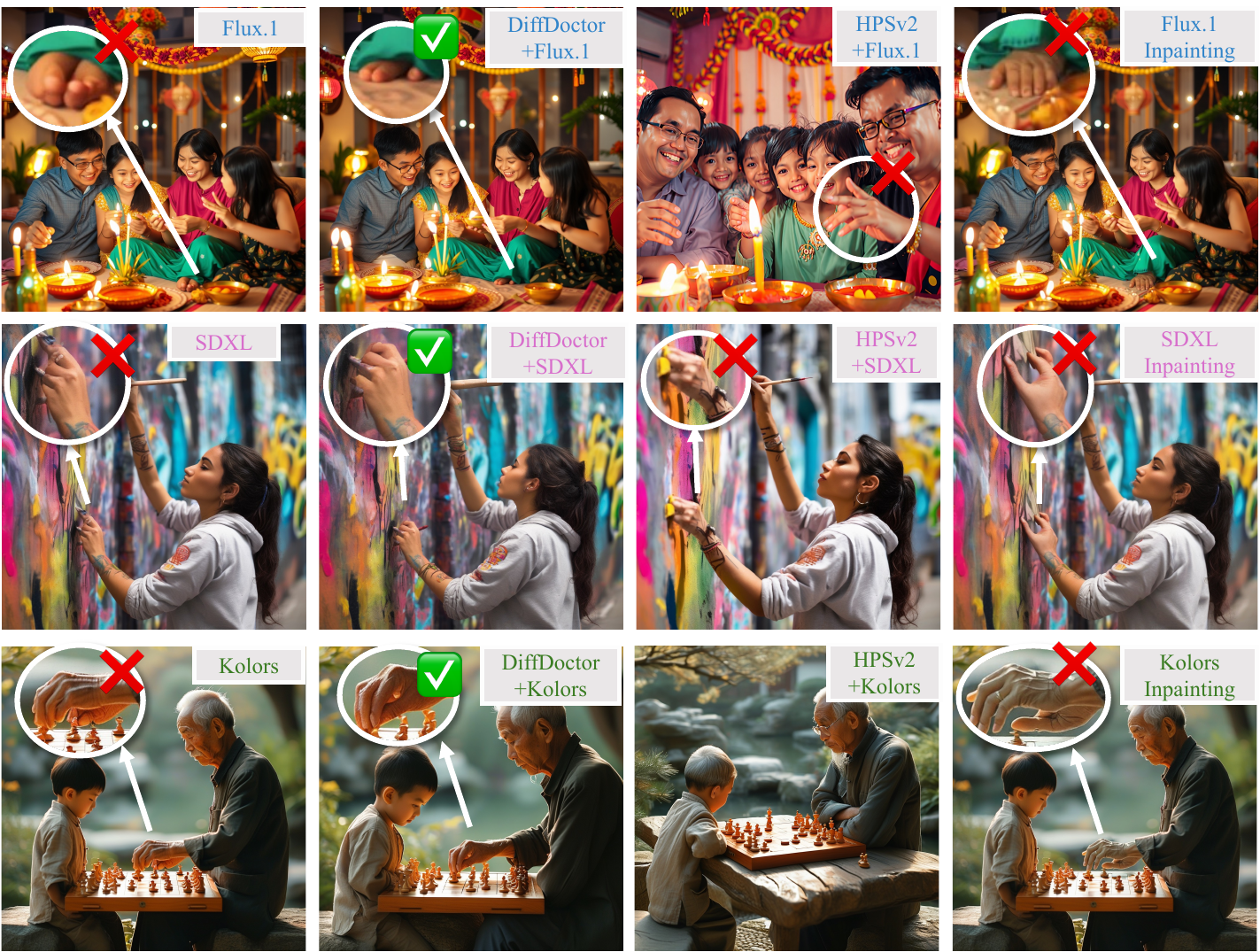} 
\vspace{-8mm}
\caption{\textbf{Qualitative comparisons} on FLUX.1, SDXL, Kolors. We compare DiffDoctor with using HPSv2 to optimize diffusion models and in-painting methods that redraw artifacts using masks.}
\label{fig: comp_rl}
\vspace{-5mm}
\end{figure}

 \begin{table}[!t]
      \centering

      \caption{\textbf{Quantitative comparisons}.  
     We compare metrics before and after \methodname\xspace, with other methods aiming to improve synthesis quality on FLUX.1, SDXL, and Kolors.
    }
    \vspace{-3mm}
    \resizebox{!}{1.9cm}{
    \begin{tabular}{lcccl}
    \toprule
    Method & Mean Artifact Freq.~($\downarrow$)  & ImageReward $\uparrow$ & CLIP-T~($\uparrow$)\\
    \midrule
    FLUX.1 & 82.66\%$\pm$3.40\%   & 1.179$\pm$0.019 & 35.463$\pm$0.047 \\
    FLUX.1 + Hands XL  & 60.54\%$\pm$0.96\% & 0.961$\pm$0.061  & 33.528$\pm$0.290 \\
    FLUX.1 + HPSv2 & 80.67\%$\pm$2.49\%   & 1.022$\pm$0.010 & 35.037$\pm$0.076 \\
    \textbf{FLUX.1 + DiffDoctor}  & 22.00\%$\pm$3.74\%   & 1.183$\pm$0.024 & 35.611$\pm$0.036 \\
    \midrule
    SDXL  & 55.33\%$\pm$0.94\%  & 0.974 $\pm$0.007 & 36.211$\pm$0.066 \\ 
    SDXL + HPSv2 & 81.19\%$\pm$1.83\%   & 0.976$\pm$0.077 & 35.037$\pm$0.154 \\
    \textbf{SDXL + DiffDoctor}   & 27.50\%$\pm$2.18\%  & 1.008 $\pm$ 0.015 & 
    36.217$\pm$0.017 \\
    \midrule
    Kolors   & 65.31\%$\pm$0.94\% & 0.823$\pm$0.039 & 34.251$\pm$0.230 \\
    Kolors + HPSv2 & 60.38\%$\pm$2.86\%   & 0.846$\pm$0.024 & 32.798$\pm$0.124 \\
    \textbf{Kolors + DiffDoctor}  & 29.33\%$\pm$3.39\%  & 0.824$\pm$0.026 & 34.424$\pm$0.124 \\
    \bottomrule
    \end{tabular}
    }
    \label{tab: quan_comp}
     \vspace{-5mm}
    \end{table}

\subsection{Comparisons on Treating}

We compare \methodname\xspace with methods aiming at improving human preference and fixing artifacts on FLUX.1, SDXL, and Kolors. We use HPSv2~\cite{wu2023human} to measure human preference and tune the diffusion model to maximize it using differentiable rewards~\cite{clark2024directly}. We also use different inpainting models to redraw the artifact areas, where user-provided masks are required, so we only compare them qualitatively.
We also include Hands XL~\cite{handsxl}, a LoRA that aims at solving hand artifacts. It only supports FLUX.1, so we only show its quantitative results (qualitative results are in the appendix).

\noindent\textbf{Qualitative comparisons for treating.}
 In \cref{fig: comp_rl}, HPSv2 doesn't significantly improve artifacts and greatly changes the image layouts. Inpainting-based methods create unnatural boundaries for masked regions and are still prone to generating defects after inpainting the problematic areas. Whereas \methodname\xspace reduces the artifacts naturally while only slightly impacting the image output after training.

\noindent\textbf{Quantitative comparisons for treating.}
In \cref{tab: quan_comp}, \methodname\xspace significantly lowers the artifact frequencies after training.
Though focusing on local artifacts, \methodname\xspace still improves two global metrics, ImageReward and CLIP-T scores, for all three backbones, showing that it keeps and even improves the image quality and image-text consistency. 
Whereas, Hands XL can reduce the artifact frequency to some extent but is not comparable to \methodname\xspace, for it only solves hands. HPSv2 only optimizes a global human preference score, so the artifact frequency remains almost unchanged for FLUX.1 and Kolors and is even higher on SDXL. 
Though HPSv2 boosts the ImageReward score on Kolors, it significantly lowers the CLIP-T score for all backbones, showing a sacrifice for text-image consistency.

User studies in \cref{tab: userstudy} further prove that \methodname\xspace successfully reduces artifacts while keeping image quality. The winning rates on image general quality are almost the same, but after treating, the winning rates on artifacts significantly outperform all the backbone model baselines.

\subsection{Qualitative Demonstrations}
\noindent\textbf{\methodname\xspace on FLUX.1.} We diagnose then treat FLUX.1 in \cref{fig: reflection_exp_FLUX}. All images are synthesized on unseen prompts and the same seeds for corresponding images at evaluation time, and each image pair shows images synthesized before and after treating. 
\methodname\xspace successfully generalizes to unseen prompts, where different artifacts (\textit{e.g.}, watermarks, distorted words, extra limbs, distorted hands, abnormal fingers, blurry head) that are of similar types in training time are suppressed in evaluation time, even when the scenarios are rich in details and complex patterns. 
We also observe that treating the model doesn't significantly change the layout of the images synthesized on the same prompt and seed in evaluation time. 
Pixel-aware treating seems like `altering' the critical areas precisely to make the model avoid artifacts for these areas, and other regions that are considered low artifact confidence areas are not greatly modified. 

\begin{figure}[!t]
  \centering
  \resizebox{1\linewidth}{!}{
      \includegraphics[width=0.4\linewidth]{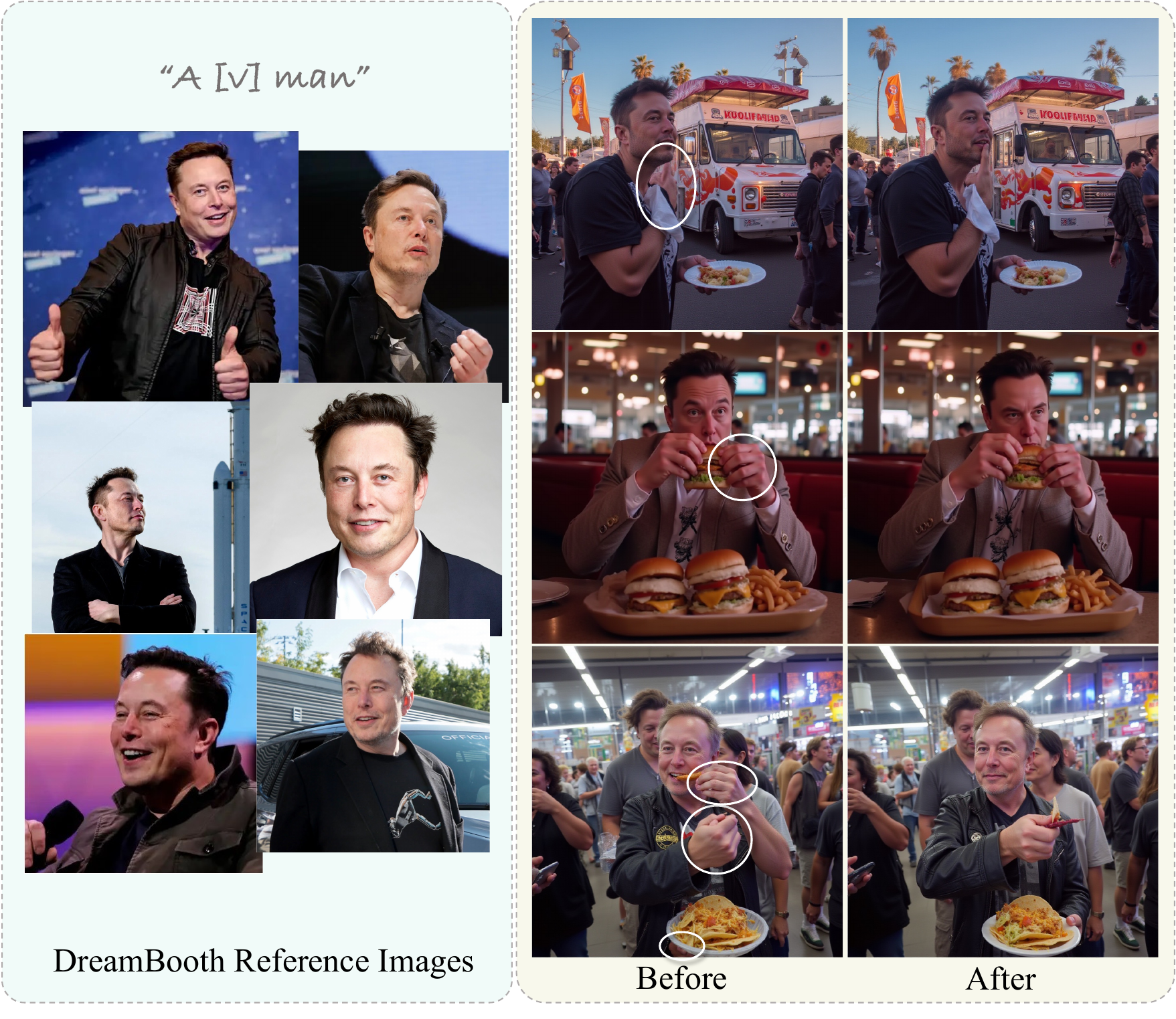}
      }
   \vspace{-7mm}
  \caption{\textbf{\method on DreamBooth.} We perform \methodname\xspace on the instance prompt and evaluate other various prompts.
  }
  \label{fig: dreambooth} 
  \vspace{-4mm}
\end{figure}

\begin{table}[t]
      \centering
      \caption{\textbf{User studies}. 
  We do user studies on 24 people, making them select a winner between two images (the same diffusion model but before and after treating) based on quality or artifacts.
  }
  \vspace{-3mm}
\resizebox{!}{1.55cm}{
  \begin{tabular}{lccl}
    \toprule
    Method & Quality Win~($\uparrow$)  & Artifact Win~($\uparrow$)\\
    \midrule
    FLUX.1 & 48.86\%   & 36.36\% \\
    \textbf{FLUX.1 + DiffDoctor}  & 51.14\%  & 63.64\% \\
    \midrule
    SDXL  & 51.89\%  & 31.82\% \\ 
    \textbf{SDXL + DiffDoctor}  & 48.11\%  & 68.18\% \\
    \midrule
    Kolors   & 49.24\% & 39.02\% \\
    \textbf{Kolors + DiffDoctor}  & 50.76\%  & 60.98\% \\
  \bottomrule
  \end{tabular}
  }

   \vspace{-5mm}
  \label{tab: userstudy}
  \end{table}

\noindent\textbf{\methodname\xspace on DreamBooth}.
In addition to T2I base models, \methodname\xspace could also benefit downstream tasks where artifacts occur.
In \cref{fig: dreambooth}, we verified its effectiveness on customized generation.   
Specifically, we train DreamBooth~\cite{ruiz2023dreambooth} on FLUX.1 Dev, with an inference step of 10.
Then we further optimize these LoRA layers using \methodname\xspace based on the instance prompt (\textit{e.g.}, ``A [v] man") and class prompts (\textit{e.g.}, ``A  man sits in the park").
Afterward, we evaluate various prompts describing the instance
(\textit{e.g.}, ``A [v] man eats hamburgers").
Results show that treating DreamBooth successfully reduces artifacts on unseen prompts while preserving the identity.
\section{Conclusion and Discussion}
\label{sec:conclusion}

We present \method, which, to the best of our knowledge, is the first approach to use pixel-level feedback for tuning diffusion models. \methodname\xspace diagnoses artifacts in synthesized images and treats the diffusion model to make them try to avoid generating artifacts in the future.
To diagnose pixel-level artifacts, we train a robust artifact detector by balancing the distribution and scaling up in a human-in-the-loop manner.  To treat the diffusion model, we supervise the diffusion model with the artifact detector.
Experiments show that, tuned on limited prompts, \methodname\xspace effectively reduces the occurrence of artifacts of similar kinds while maintaining image quality on unseen prompts, validated by qualitative and quantitative results.
It is also applicable to other methods such as DreamBooth.
\methodname\xspace serves as a new post-training paradigm to improve the generation ability of pre-trained image diffusion models.

\noindent\textbf{Limitations.}
Despite impressive results, \methodname\xspace currently cannot deal with semantically challenging artifacts that are extremely complicated due to the lack of a strong artifact detector capable of complex reasoning to find out (\textit{i.e.,} objects that are against physical laws or common knowledge).
We further discuss this limitation in the appendix and leave this topic for future studies. 

\clearpage
{
\small
\renewcommand\UrlFont{\color{Gray}\ttfamily}
\bibliographystyle{ieeenat_fullname}
\bibliography{main}
}

\end{document}